# Machine Learning Classifications of Coronary Artery Disease

[1]Ali Bou Nassif, [1]Omar Mahdi, [1]Qassim Nasir, [2]Manar Abu Talib

[1]Department of Electrical and Computer Engineering, University of Sharjah

[2]Department of Computer Science, University of Sharjah Sharjah, UAE

{anassif, U16105933, nasir, mtalib}@sharjah.ac.ae

Mohammad Azzeh
Department of Software Engineering
Applied Science Private University
Amman, Jordan, POBOX 166
m.y.azzeh@asu.edu.jo

*Abstract*—Coronary Artery Disease (CAD) is one of the leading causes of death worldwide, and so it is very important to correctly diagnose patients with the disease. For medical diagnosis, machine learning is a useful tool; however features and algorithms must be carefully selected to get accurate classification. To this effect, three feature selection methods have been used on 13 input features from the Cleveland dataset with 297 entries, and 7 were selected. The selected features were used to train three different classifiers, which are SVM, Naïve Bayes and KNN using 10-fold cross-validation. The resulting models evaluated using Accuracy, Recall, Specificity and Precision. It is found that the Naïve Bayes classifier performs the best on this dataset and features, outperforming or matching SVM and KNN in all the four evaluation parameters used and achieving an accuracy of 84%.

## I. INTRODUCTION

Cardiovascular diseases are one of the leading causes of death worldwide, causing around a third of deaths every year. Out of all cardiovascular diseases, coronary artery disease is the cause of most deaths [1][2]. As such, there is great interest in diagnosing, treating and preventing this disease.

To this effect, machine learning algorithms have been greatly utilized to try and create models that help in the detection of this disease by building classification algorithms that predict whether a person has CAD or not based on medical databases.

The "heart-disease directory" contains datasets with 14 commonly used features that are used to train machine learning models to classify whether a person has the disease or not. In this paper, three different feature selection techniques are used to select the most relevant features for classification, the parameters of three classification algorithms are tuned and then evaluated and compared.

## II. TECHNICAL BACKGROUND

### A. Coronary Artery Disease

CAD is a heart disease where the arteries of the heart, specifically the coronary arteries, become narrower limiting the amount of blood that can go through which reduces the amount of oxygen delivered to the heart and further stressing the heart as it has to work harder to pump blood through increasingly narrower passages. Once an artery is completely blocked a heart attack occurs.

This blockage of the arteries usually happens by fat deposits building up in the artery and therefore slowly narrowing it [3]. A major cause of the disease is the lack of exercise, and it has been shown that regular can greatly help the prevention and rehabilitation from CAD [4].

### B. Support Vector Machine

A Support Vector Machine (SVM) is a supervised machine learning algorithm that is mainly used as a classifier [5], although it can also be adjusted to work for regression problems [6].

SVMs work by separating data points using hyperplanes. Where a data point is an $n$ dimensional feature vector, the hyperplane is the geometric shape that occupies $n-1$ dimensions. The hyperplane is used as the decision boundary between the two data classes, where a new data point will be classified according to which side of the hyperplane it lies on.

In the simplest case where the data is linearly separable, two parallel hyperplanes are used to separate the classes such that the distance between the hyperplanes is maximum, which minimizes the classification error. The hyperplane that lies in the middle point between them forms the decision boundary for classification, and the observations that lie on the hyperplanes are known as the support vectors.

However, if the data is not linearly separable, kernels can be used to produce non-linear classifiers [5] by transforming the features into a higher dimensional space where they become linearly separable by a hyperplane.

*C. Naïve Bayes*

Naïve Bayes classifiers are based on applying Bayes' probability theorem to classify a given data point. These classifiers work under the assumption that all the features are completely independent of each other, even if this is an oversimplification, therefore called 'naïve'.

Bayes' rule gives the probability of an event given some relation with another variable, where the simple form is [7]:

$$P(C_i | x) = \frac{P(x | C_i) P(C_i)}{P(x)}, P(x) \neq 0 \quad (1)$$

In a classification scenario: $C_i$ is the $i^{th}$ class, $x$ is the feature vector of independent variables, $P(C_i | x)$ is the probability of $x$ being of class $C_i$ and the last term $P(x | C_i)$ the probability of getting $x$ given that the class is $C_i$. An observation, $x$, will be classified according to which $C_i$ gives the highest $P(C_i | x)$ for it.

*D. K-Nearest Neighbors*

The KNN algorithm is a relatively simple supervised machine learning algorithm that can be used for both classification [8] and regression [9] tasks. The KNN algorithm classifies a given data point according to a voting procedure that is based on the class types of the nearest $k$ examples in multidimensional space.

In its simplest form, the KNN technique classifies a data point purely on which class type dominates for the $k$ closest points. For example, given $k = 3$ and an object to classify $x$, the algorithm will first find the closest three data points to $x$ via a distance measure, such as Euclidean distance, which is defined as:

$$D(x, p) = \sqrt{\sum_{i=1}^{n} (x_i - p_i)^2} \quad (2)$$

Where $D(x, p)$ is the distance between the object to be classified $x$ and a point $p$, and $n$ is the length of the feature vector. Once the closest $k$ points are identified, the class type that occurs the most within the closest $k$ points is chosen as the class of $x$. Note that $k$ is usually an odd number to avoid tie situations.

However, more sophisticated techniques to decide the class of a given data point have been developed, such as using weighted distances instead of pure ones [10], which can improve the classification accuracy of the KNN algorithm and its speed [11].

III. DATASET

The "heart-disease directory" available in the "UCI Machine Learning Repository" [12] contains a total of 4 databases related to heart disease diagnosis, where each dataset contains the same 75 features and one categorical output, however only 13 of the input variables have been used in research so far [13][14][15].

Three of the datasets have only a few (2-5) entries that are fully valid, while the other entries have missing values, therefore, the "Cleveland Clinic Foundation" dataset was chosen as it only has 6 entries with missing data, leaving 297 rows to be used.

It should be noted that some of the independent variables have been renamed to make them easier to understand, with these changes marked by brackets.

*A. Features*

Explanation of each input feature and its possible values, where applicable, are shown below.

1. Age
2. Sex
   - 0: Female
   - 1: Male
3. Cp (Chest Pain)
   - 1: typical angina
   - 2: atypical angina
   - 3: non-anginal pain
   - 4: asymptomatic
4. Trestbps (Restbp); Resting blood pressure (in mm Hg on admission to the hospital)
5. Chol; serum cholesterol in mg/dl
6. fbs; Is fasting blood sugar > 120 mg/dl
   - 0: False
   - 1: True
7. RestECG; Resting electrocardiographic results
   - 0: Normal
   - 1: Having ST-T wave abnormality (T wave inversions and/or ST elevation or depression of > 0.05 mV)
   - 2: Showing probable or definite left ventricular hypertrophy by Estes' criteria
8. thalach (Max Heart); Maximum heart rate achieved
9. ExAng; Exercise induced angina
   - 0: No
   - 1: Yes
10. OldPeak; ST depression induced by exercise relative to rest
11. Slope; The slope of the peak exercise ST segment
    - 1: Up sloping
    - 2: Flat

- 3: Down sloping

12. Ca (MajorVessels); Number of major vessels (0-3) colored by fluoroscopy

13. Thal;

- 3: Normal
- 6: Fixed defect
- 7: Reversible defect

14. Num (Positive); The final diagnosis of heart disease (angiographic disease status)

- 0: < 50% diameter narrowing
- 1: > 50% diameter narrowing.

## IV. RELATED WORK

Several previous work have used the datasets of the heart-disease directory with machine learning algorithms to predict the presence or absence of CAD. Of the four datasets available in the heart-disease directory, the Cleveland dataset is generally the most utilized due to the very low number of missing values compared to the other datasets.

Work done in [16] compared Naïve Bayes, J48 decision trees and an artificial neural network(ANN) and saw that the Naïve Bayes algorithm performed the best. Setiawan et al. [17], [18] also used the heart-disease directory datasets but they both employed fuzzy logic techniques, with [17] testing KNN and decision trees, among others.

Muthukaruppan et al. [19] used particle swarm optimization with a fuzzy expert system to classify CAD and achieve high accuracy, while [20] used an evolutionary fuzzy expert system on the Cleveland dataset to carry out prediction. Both [19] and [20] used decision trees to build the fuzzy rule base.

However, none of these previous work have used several feature selection techniques to select the training features from the 14 that are commonly used, and the classification algorithms used in each did not carry out direct comparisons between the three different classification techniques discussed in this paper.

The results presented in this paper show how correctly selecting the features and model parameters provided an improvement in all the evaluation parameters relative to previous works that used SVM, KNN and Naïve Bayes. This can be seen by the review made in [18] between the performance of these algorithms in previous works.

## V. METHODOLOGY

### A. Feature Selection

Although the number of attributes is already small, it is always beneficial to ensure that all the features used are relevant and to remove any redundant variables, which can help improve training time and reduce overfitting [21].

In this paper, feature selection was done through the 'Waikato Environment for Knowledge Analysis' (Weka) [22] on the Cleveland dataset by using the following methods:

1. Information Gain evaluator with Ranker search
2. Correlation evaluator with Ranker search
3. Classifier Subset evaluator on Naïve Bayes with Best First search

The results for the methods with Ranker search are shown below in Fig. 1 and the Classifier Subset results, using 10-fold cross-validation on Naïve Bayes, with a seed of 1 are shown in Fig. 2.

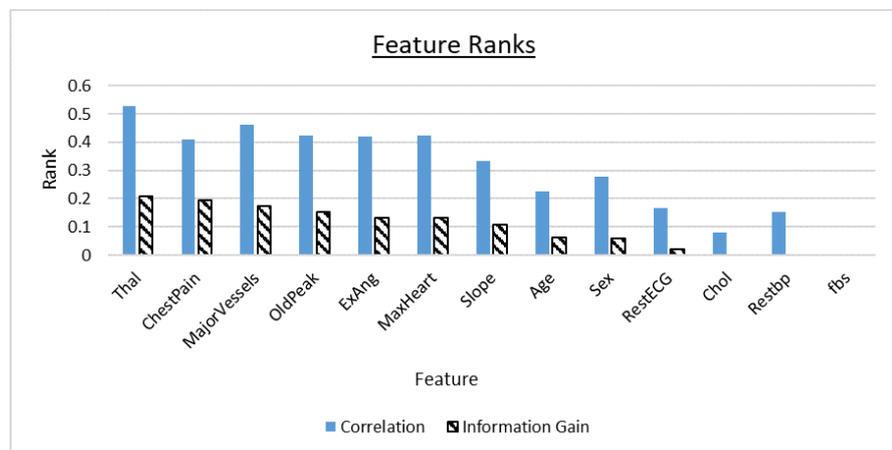

Fig. 1. Feature ranks using Information Gain and Correlation

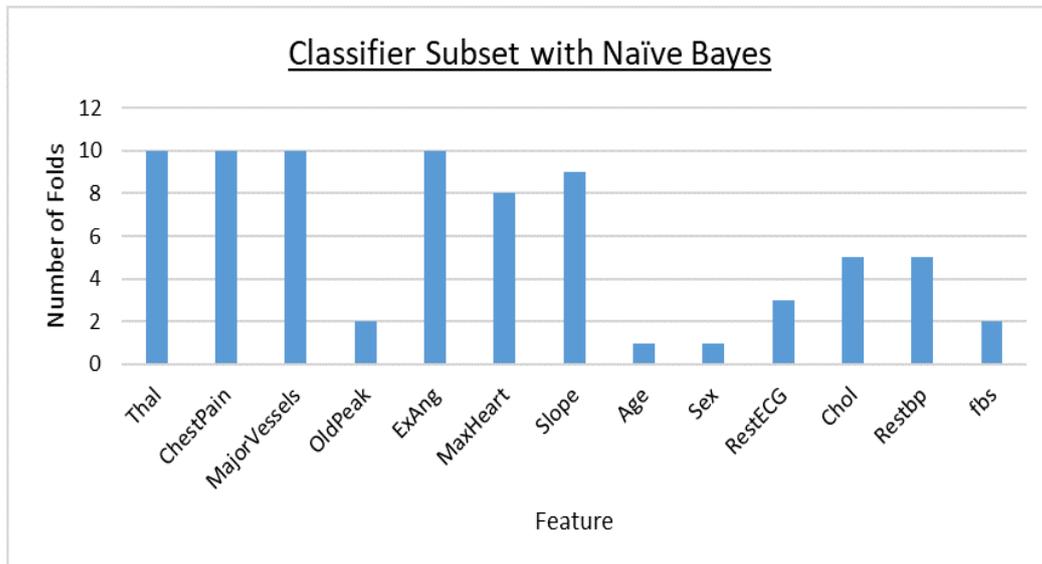

Fig. 2. Naïve Bayes Classifier Subset results

All the three methods tested paint a very similar picture in regards to the features that affect the possibility of testing positive for a coronary artery disease. The methods show that Age, Sex, Chol, fbs, Restbp and RestECG have a smaller effect on the outcome of the classification, and so they were removed from the dataset.

Based on this, the final input variables are:

1. Cp (Chest Pain)
2. Thalach (MaxHeart)
3. ExAng
4. OldPeak
5. Slope
6. Ca (MajorVessels)
7. Thal

*B. Model Design*

The 'R' programming language was used to carry out all further model training, testing and evaluation. All the variables were taken as numeric, except for the output variable, which was nominal. R's 'Caret' package was used to produce the models using 10-fold cross-validation with a randomization seed = 2018.

Parameter tuning was also done to find the best parameters for each algorithm tested. The parameters that produced the model with the highest observed accuracy were chosen, and then that model was trained on all the data to produce the final model ready to be used with new data.

Table 1 shows the parameters tested for each algorithm and the final parameters chosen.

**Table 1.** Tested parameter values and final models

| Algorithm | Caret package method | Parameters Tested | Best Parameters |
| --- | --- | --- | --- |
| **SVM** | "svmRadial" | $C$ = 0.25, 0.5 and 1.0 $Sigma$ = 0.1268408 | $C$ = 0.25 |
| **KNN** | "knn" | $K$ = 5, 7 and 9 | $K$ = 5 |
| **Naïve Bayes** | "naive_bayes" | $Kernel$ = True, False<br>Laplace Correction = 0<br>Bandwidth Adjustment = 1 | $Kernel$ = False |

## VI. RESULTS AND DISCUSSION

With the models finalized, statistical evaluation parameters to measure the performance of each model can now be calculated, and comparisons between the different models can therefore be made.

The parameters used to evaluate each model are:

$$Accuracy = \frac{TP+TN}{TP+FP+TN+FN} \quad (3)$$

$$Recall = \frac{TP}{TP+FN} \quad (4)$$

$$Specificity = \frac{TN}{TN+FP} \quad (5)$$

$$Precision = \frac{TP}{TP+FP} \quad (6)$$

Where TP is the number of correctly classified coronary artery disease cases and FP are normal cases wrongly classified as positive. TN are patients without the disease correctly classified, and FN are patients that have coronary artery disease but were misclassified as disease free.

Observed accuracy is how many entries were correctly classified. Recall signifies how many of the total positive cases, in this case a coronary heart disease diagnosis, were correctly classified, while Specificity does the same as recall but for negative samples. Precision specifies how accurately the positive classification was, irrelevant of the positive cases that were misclassified as negative.

The outcomes for each parameter are summarized in Fig. 3.

In such a medical diagnosis situation, it is very important to correctly classify as much of the positive cases as possible since incorrectly diagnosing a sick person as disease free might be fatal, while marking a healthy person as sick would likely cause less damage. Therefore, the aim is to get as little false negatives as possible, and higher true positives.

In terms of the performance measures, this means that the best classifier would ideally have high recall and precision. From Fig. 3 it can be seen that both the SVM and Naïve Bayes classifiers have equal recall, however Naïve Bayes has a slightly better precision. Not only does the Bayesian classifier perform better in these two important measures, but it has better performance in the other two parameters.

The Naïve Bayes classifier also performs surprisingly better at classifying healthy people compared to both SVM and KNN, even though it has the same recall as SVM. KNN on the other hand gives the worst results in all four of the tested performance measures.

## VII. CONCLUSION

This paper compares three classifiers named SVM, KNN and Naïve Bayes for Coronary Artery Disease using Accuracy, Recall, Specificity and Precision performance criteria on the Cleveland dataset. Feature selection techniques have been used to select the most relevant features from the dataset. Results show that the Naïve Bayes classifier outperforms SVM and KNN on the Cleveland dataset.

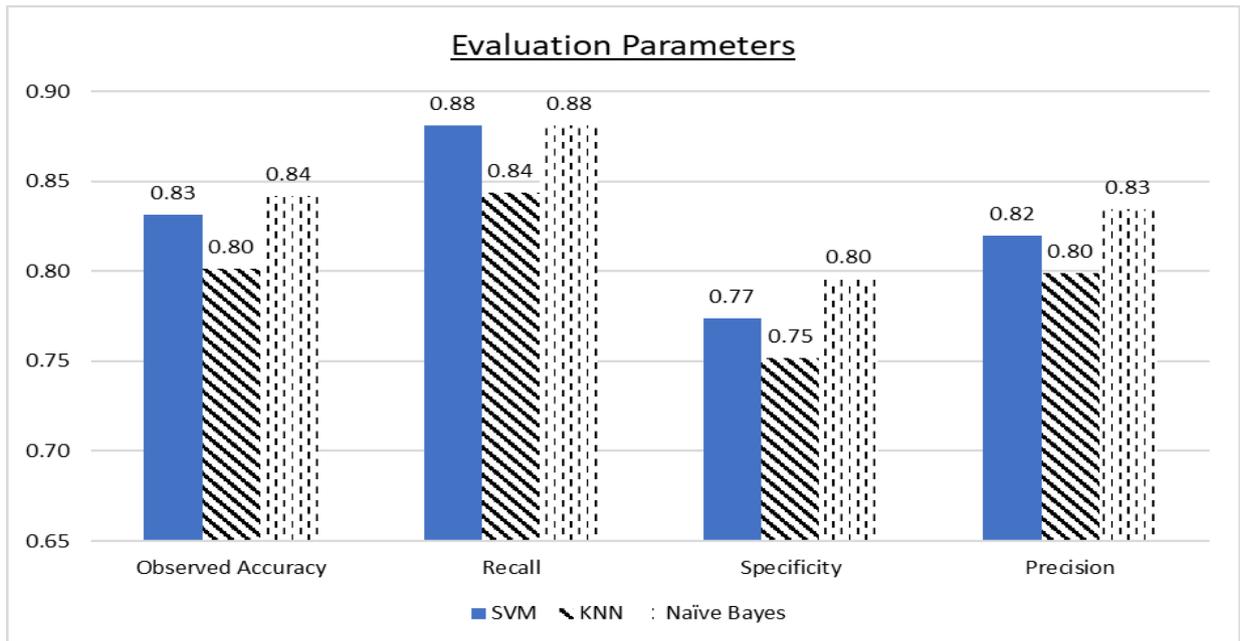

Fig. 3. Evaluation parameters for each classifier

ACKNOWLEDGEMENT

The authors would like to thank the University of Sharjah and the Open UAE Research and Development research group for their support in conducting this research